\documentclass[journal]{IEEEtran}
\usepackage{graphicx}
\usepackage{multirow}
\usepackage{longtable}
\usepackage{rotating}
\usepackage{color}
\usepackage{makecell}
\usepackage{hyperref}
\usepackage{cite}
\usepackage{lineno}
\graphicspath{{figures/}}

\ifCLASSINFOpdf
\else
\fi
\begin{document}
%
\title{DenseFuse: A Fusion Approach to Infrared and Visible Images}
%
%
%

\author{Hui Li and Xiao-Jun Wu
\thanks{This work was supported by the National Natural Science Foundation of China(Grant No.61672265, U1836218, 61876072), and the 111 Project of Ministry of Education of China (Grant No.B12018). \emph{(Corresponding author: Xiao-Jun Wu)}}
\thanks{The authors are with the School of Internet of Things Engineering, Jiangnan University, Wuxi 214122, China.(e-mail:hui\_li\_jnu@163.com; xiaojun\_wu\_jnu@163.com).}
\thanks{This paper has supplementary downloadable material available at \url{http://ieeexplore.ieee.org}, provided by the author. The material includes supplemental material. Contact xiaojun\_wu\_jnu@163.com for further questions about this work. Code and pre-trained models are available at \url{https://github.com/hli1221/imagefusion_densefuse}.}
}

\markboth{}%
{Shell \MakeLowercase{\textit{et al.}}: DenseFuse: A Fusion Approach to Infrared and Visible Images}

\maketitle

\begin{abstract}
In this paper, we present a novel deep learning architecture for infrared and visible images fusion problem. In contrast to conventional convolutional networks, our encoding network is combined by convolutional layers, fusion layer and dense block in which the output of each layer is connected to every other layer. We attempt to use this architecture to get more useful features from source images in encoding process. And two fusion layers(fusion strategies) are designed to fuse these features. Finally, the fused image is reconstructed by decoder. Compared with existing fusion methods, the proposed fusion method achieves state-of-the-art performance in objective and subjective assessment. 
\end{abstract}


\IEEEpeerreviewmaketitle

\section{Introduction}

\IEEEPARstart{T}{he} infrared and visible image fusion task is an important problem in image processing field. It attempts to extract salient features from source images, then these features are integrated into a single image by appropriate fusion method\cite{1}. For decades, these fusion methods achieve extrodinary fusion performance and are widely used in many applications, like video surveillance and military applications.

As we all know, many signal processing methods have been applied to the image fusion task to extract image salient features, such as muli-scale decomposition-based methods\cite{2,3,4,5,6,7}. Firstly, the salient features are extracted by image decomposition methods. Then, an appropriate fusion strategy is utilized to obtain the final fused image.

In recent years, the representation learning-based methods have also attracted great attention. In sparse domain, many fusion methods are proposed, such as sparse representation(SR) and Histogram of Oriented Gradients(HOG)-based fusion method\cite{8}, joint sparse representation(JSR)\cite{9} and co-sparse representation\cite{10}. In low-rank domain, Li et al.\cite{11} proposed a low-rank representation(LRR)-based fusion method. They use LRR instead of SR to extract features, then $l_1$-norm and the max selection strategy are used to reconstruct the fused image.

With the rise of deep learning, many fusion methods based on deep learning are proposed. The convolutional neural network(CNN) is used to obtain the image features and reconstruct the fused image\cite{12,13}. In these CNN-based fusion methods, only the last layer results are used as the image features and this operation will lose a lot of useful information which is obtained by middle layers. We think these are important for fusion method.

In order to solve this problem, in our paper, we propose a novel deep learning architecture which is constructed by encoding network and decoding network. We use encoding network to extract image features and the fused image is obtained by decoding network. The encoding network is constructed by convolutional layer and dense block\cite{14} in which the output of each layer is used as the input of next layer. So in our deep learning architecture, the results of each layer in encoding network are utilized to construct feature maps. Finally, the fused image will be reconstructed by fusion strategy and decoding network which includes four CNN layers.

Our paper is structured as follows. In Section\ref{sec:related}, we briefly review related works. In Section\ref{sec:proposed}, the proposed fusion method is introducted in detail. The experimental results are shown in Section\ref{sec:experimrnt}. The conclusion of our paper with discussion are presented in section \ref{sec:con}.

\begin{figure*}[!ht]
\centering
\includegraphics[width=\linewidth]{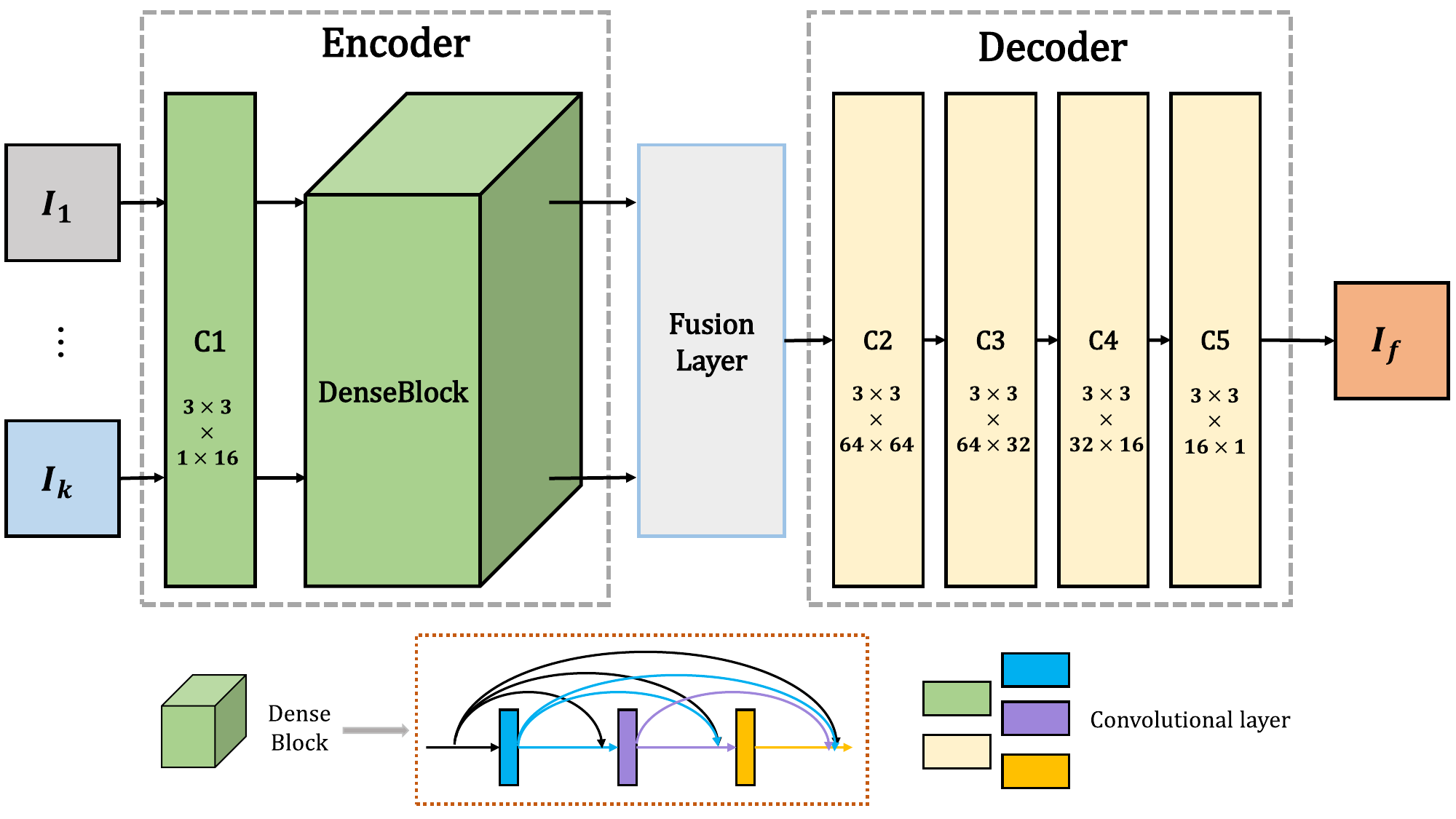}
\caption{The architecture of proposed method.}
\label{fig:architecture}
\end{figure*}

\section{Related Works}
\label{sec:related}
Many fusion algorithms have been proposed in the last two years, especially based on deep learning. Unlike multi-scale decomposition-based methods and representation learning-based methods, the deep learning-based algorithms use lot of images to train the network and these networks are used to obtain the salient features.

In 2016, Yu Liu et al.\cite{12} proposed a fusion method based on convolutional sparse representation(CSR). The CSR is different from CNN-based methods, but this algorithm is still deep learning-based algorithm, because it also extracts the deep features. In this method, authors use source images to learn several dictionaries which have different scales and employ CSR to extract multi-layer features, then fused image is generated by these features. In 2017, Yu Liu et al.\cite{13} also presented a CNN-based fusion method for multi-focus image fusion task. The image patches which contain different blur versions of the input image are used to train the network and use it to get a decision map. Then, the fused image is obtained by using the decision map and the source images. However, this method is only suitable for multi-focus image fusion.

In ICCV 2017, Prabhakar et al.\cite{15} performed a CNN-based approach for exposure fusion problem. They proposed a simple CNN-based architecture which contains two CNN layers in encoding network and three CNN layers in decoding network. Encoding network has siamese network architecture and the weights are tied. Two input images are encoded by this network. Then, two feature map sequences are obtained and they are fused by addition strategy. The final fused image is reconstructed by three CNN layers which is called decoding network. Although this method achieves better performance, it still suffers from two main drawbacks: 1) The network architecture is too simple and the salient features may not be extracted properly; 2) These methods just use the result which is calculated by the last layers in encoding network and useful information obtained by middle layers will be lost, this phenomenon will get worse when the network is deeper.

To overcome these drawbacks, we propose a novel deep learning architecture based on CNN layers and dense block. In our network, we use infrared and visible image pairs as inputs for our method. And in dense block, their feature maps which are obtained by each layer in encoding network are cascaded as the input of the next layer.

In traditional CNN based network, with the increase of network depth, a degradation problem\cite{15} has been exposed and the information which is extracted by middle layers is not used thoroughly. To address the degradation problem, He et al.\cite{16} introduced a deep residual learning framework. To further improve the information flow between layers, Huang et al.\cite{14} proposed a novel architecture with dense block in which direct connections from any layer to all the subsequent layers are used. Dense block architecture has three advantages: 1) this architecture can preserve as much information as possible; 2) this model can improve flow of information and gradients through the network, which makes network be trained easily; and 3) the dense connections have a regularizing effect, which reduces overfitting on tasks.

Based on these observations, we incorporate dense block in our encoding network, which is the origin of our proposed name: Densefuse. With this operation, our network can preserve more useful information from middle layers and easy to train. We will introduce our fusion algorithm in detail in Section \ref{sec:proposed}.

\begin{figure*}[!ht]
\centering
\includegraphics[width=\linewidth]{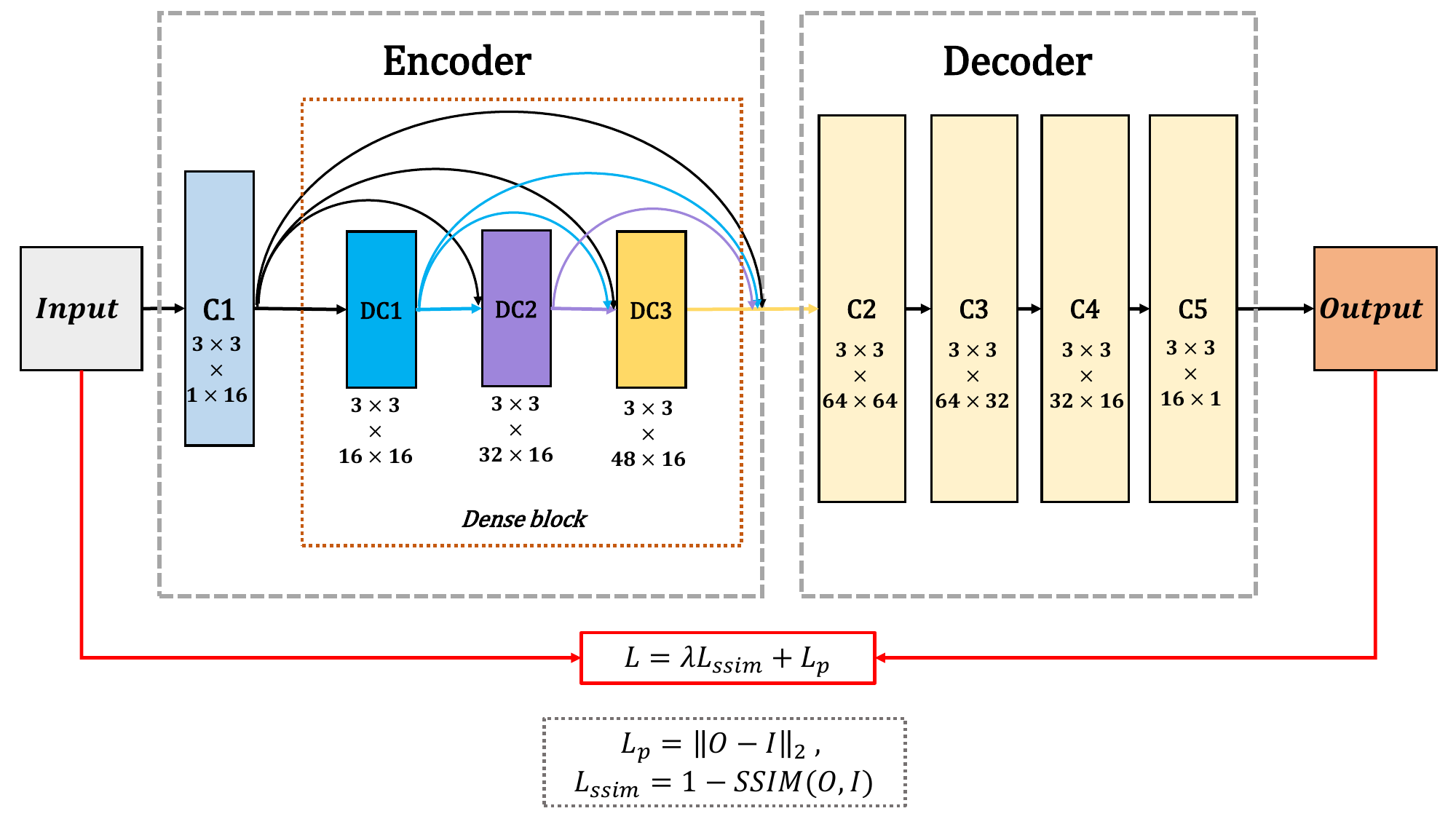}
\caption{The framework of training process.}
\label{fig:train}
\end{figure*}

\section{Proposed Fusion Method}
\label{sec:proposed}
In this section, the proposed deep learning-based fusion method is introduced in detail. With a span of last 5 years, CNN gains great success in image processing field. It is also the footstone for our network.

The fusion strategy for color images(RGB) is the same as gray scale images in our fusion framework, so in our paper, we just consider the gray scale image fusion task. The fused results of color images are shown in Section \ref{sec:ivrgb}.

The input infrared and visible images(gray scale images) are denoted as $I_1,\cdots,I_k$, and $k\geq 2$. Note that the input images are pre-registered. And the index($1,\cdots,k$) is irrelevant with the type of input images, which means $I_i(i=1,\cdots,k)$ can be treated as infrared or visible image. We assume that input images are registered using existing algorithms \cite{31} \cite{32} \cite{33}, and in our paper, we use the method in \cite{31} to pre-process input images if they are not registered. Our network architecture has three parts: encoder, fusion layer, and decoder. The architecture of the proposed network is shown in Fig.1.

As shown in Fig.\ref{fig:architecture}, the encoder contains two parts (C1 and DenseBlock) which are utilized to extract deep features. The first layer (C1) contains $3\times3$ filters to extract rough features and the dense block (DenseBlock) contains three convolutional layers (each layer's output is cascaded as the input of the next layer) which also contain $3\times3$ filters. And in our network, the reflection mode is used to pad input image. For each convolutional layer in encoding network, the input channel number of feature maps is 16. The architecture of encoder has two advantages. First, the filter size and stride of convolutional operation are $3\times3$ and 1, respectively. With this strategy, the input image can be any size. Second, dense block architecture can preserve deep features as much as possible in encoding network and this operation can make sure all the salient features are used in fusion strategy.

We choose different fusion strategies in fusion layer and these will be introduced in Section\ref{sec:strategy}.

The decoder contains four convolutional layers ($3\times3$ filters). The output of fusion layer will be the input of decoder. We use this simple and effective architecture to reconstruct the final fused image.

\subsection{Training}
\label{sec:training}

In training phase, we just consider encoder and decoder networks(fusion layer is discarded), in which we attempt to train our encoder and decoder networks to reconstruct the input image. After the encoder and decoder weights are fixed, we use adaptive fusion strategy to fuse the deep features which are obtained by encoder. The detailed framework of our network work in training phase is shown in Fig.\ref{fig:train}, and the architecture of our network is outlined in Table \ref{tab:trainframe}.

The apparent advantage of this training strategy is that we can design appropriate fusion layer for specific fusion tasks. Furthermore, it leaves more space for further development of fusion layer.

In Fig.\ref{fig:train} and Table \ref{tab:trainframe}, C1 is convolution layer in encoder network which contains $3\times3$ filters. DC1, DC2 and DC3 are convolution layers in dense block and the output of each layer is connected to every other layer by cascade operation. The decoder consists of C2, C3, C4 and C5, which will be utilized to reconstruct the input image.
\begin{table}[ht]
\centering
\caption{\label{tab:trainframe}The architecture of training process. \emph{\textbf{Conv}} denotes the convolutional block(convolutional layer + activation); \emph{\textbf{Dense}} denotes the dense block.}
\resizebox{3.5in}{!}{
\begin{tabular}{|c|c|c|c|c|c|c|}
\hline
 & Layer & Size & Stride & \makecell[cc]{Channel\\(input)} & \makecell[cc]{Channel\\(output)} & Activation \\
\hline
\multirow{2}*{Encoder} &
Conv(C1)      & 3      & 1      & 1      & 16      & ReLu \\
~ & Dense     &        &        &        &         & \\
\hline
\multirow{4}*{Decoder} &
    Conv(C2)     & 3      & 1      & 64     & 64      & ReLu \\
~ & Conv(C3)     & 3      & 1      & 64     & 32      & ReLu \\
~ & Conv(C4)     & 3      & 1      & 32     & 16      & ReLu \\
~ & Conv(C5)     & 3      & 1      & 16     & 1       & \\
\hline
\multirow{3}*{\makecell[cc]{Dense\\(dense block)}} &
    Conv(DC1)     & 3      & 1      & 16     & 16      & ReLu \\
~ & Conv(DC2)     & 3      & 1      & 32     & 16      & ReLu \\
~ & Conv(DC3)     & 3      & 1      & 48     & 16      & ReLu \\
\hline
\end{tabular}}
\end{table}

In order to reconstruct the input image more precisely, we minimize the loss function $L$ to train our encoder and decoder,
\begin{eqnarray}\label{Eq1}
  	L = \lambda L_{ssim}+L_p
\end{eqnarray}
which is a weighted combination of pixel loss $L_p$ and structural similarity (SSIM) loss $L_{ssim}$ with the weight $\lambda$.

The pixel loss $L_p$ is calculated as,
\begin{eqnarray}\label{Eq2}
  	L_p = ||O-I||_2
\end{eqnarray}
where $O$ and $I$ indicate the output and input images, respectively. It is the Euclidean distance between the output $O$ and the input $I$.

The SSIM loss $L_{ssim}$ is obtained by Eq.\ref{Eq3},
\begin{eqnarray}\label{Eq3}
  	L_{ssim} = 1-SSIM(O,I)
\end{eqnarray}
where $SSIM(\cdot)$ represents the structural similarity operation\cite{17} and it denotes the structural similarity of two images. Because there are three orders of magnitude difference between pixel loss and SSIM loss, in training phase, so the $\lambda$ is set as 1, 10, 100 and 1000, respectively.

The aim of training phase is to train an autoencoder network(encoder, decoder) which has better feature exteraction and reconstruction ability. Due to the training data of infrared and visible images is insufficient, we use gray scale images of MS-COCO\cite{18} to train our model.

In our training phase, we use visible images to train the weights of encoder and decoder. And we train our network using MS-COCO\cite{18} as input images which contains 80000 images and all of them are resized to $256\times256$ and transformed to gray scale images. Learning rate is set as $1\times 10^{-4}$.The batch size and epochs are 2 and 4, respectively. Our method was implemented with NVIDIA GTX 1080Ti GPU. And Tensorflow is utilized as the backend for the network architecture. The analysis of training phase will be introduced in Section \ref{sec:trainingana}.

\begin{figure}[!ht]
\centering
\includegraphics[width=\linewidth]{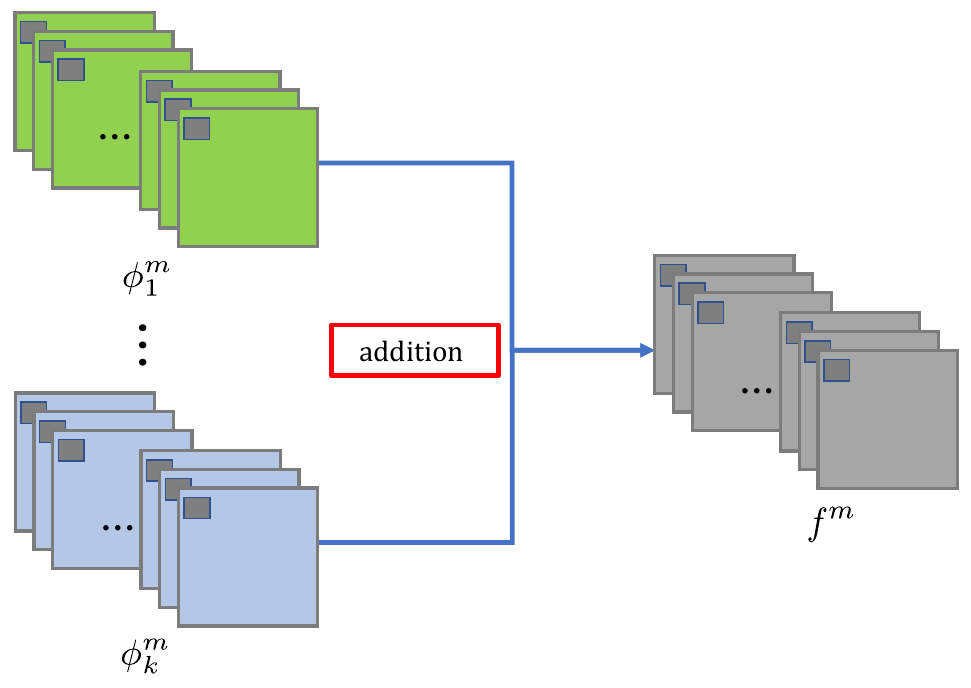}
\caption{The procedure of addition strategy.}
\label{fig:addition}
\end{figure}

\subsection{Fusion Layer(strategy)}
\label{sec:strategy}

\subsubsection{Addition Strategy}
The addition fusion strategy just like the fusion strategy in \cite{15}. And the strategy procedure is shown in Fig.\ref{fig:addition}.
As shown in Fig.\ref{fig:architecture}, once encoder and decoder networks are fixed, in testing phase, two input images are fed into encoder, respectively. We choose two fusion strategies (addition strategy and $l_1$-norm strategy) to combine salient feature maps which are obtained by encoder.

In our network, $m\in{\{1,2,\cdots,M\}}, M=64$ represents the number of feature maps. $k\geq 2$ indicates the index of feature maps which are obtained from input images.

Where $\phi_i^m(i=1,\cdots,k)$ indicates the feature maps obtained by encoder from input images, $f^m$ denotes the fused feature maps. The addition strategy is formulated by Eq.\ref{Eq4},
\begin{eqnarray}\label{Eq4}
  	f^m(x,y) = \sum_{i=1}^k \phi_i^m(x,y)
\end{eqnarray}
where $(x,y)$ denotes the corresponding position in feature maps and fused feature maps. Then $f^m$ will be the input to decoder and final fused image will be reconstructed by decoder.

\subsubsection{$l_1$-norm Strategy}
\begin{figure}[!ht]
\centering
\includegraphics[width=\linewidth]{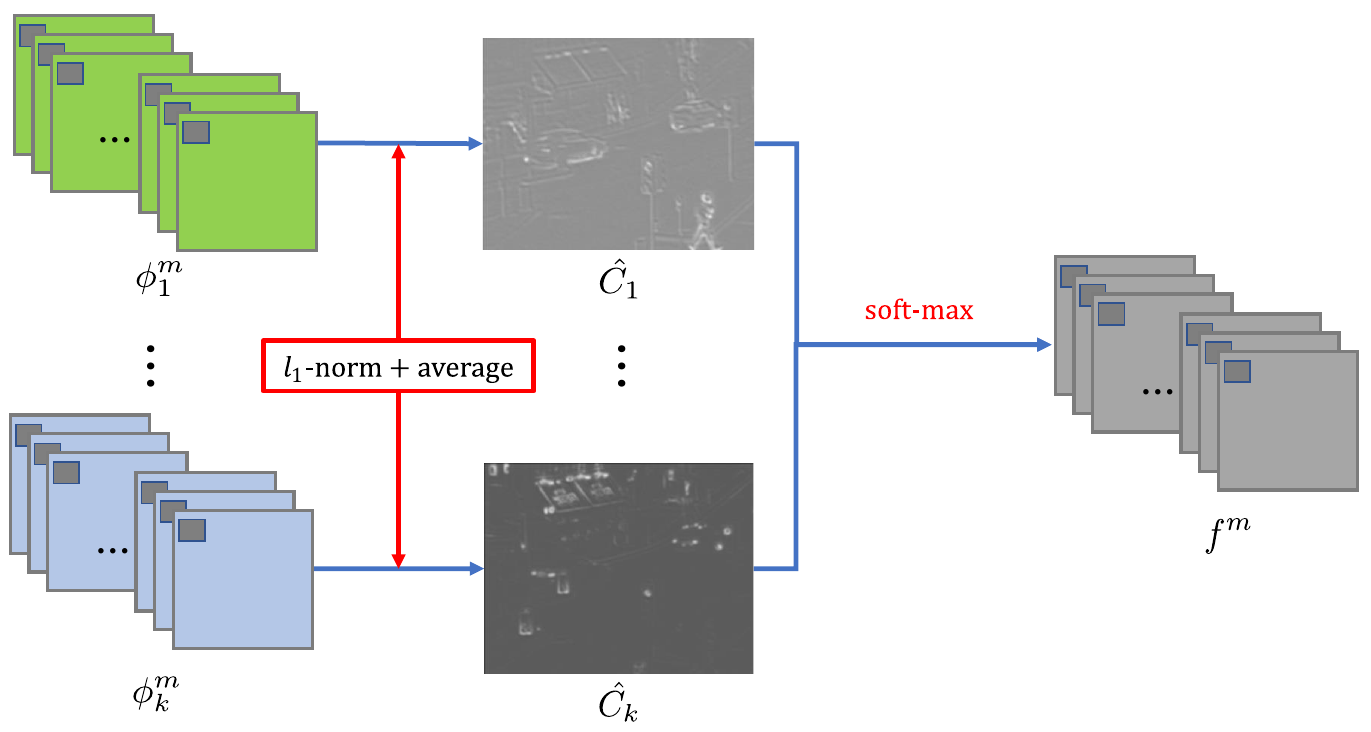}
\caption{The diagram of $l_1$-norm and soft-max strategy.}
\label{fig:l1norm}
\end{figure}

The performance of addition strategy was proved in \cite{15}. But this operation is a very rough fusion strategy for salient feature selection. We applied a new strategy which is based on $l_1$-norm and soft-max operation into our network. The diagram of this strategy is shown in Fig.\ref{fig:l1norm}.

In Fig.\ref{fig:l1norm}, the features maps are denoted by $\phi_i^m$, the activity level map $\hat{C_i}$ will be calculated by $l_1$-norm and block-based average operator, and $f^m$ still denotes the fused feature maps.

Inspired by \cite{12}, the $l_1$-norm of $\phi_i^{1:M}(x,y)$ can be the activity level measure of the feature maps. Thus, the initial activity level map $C_i$ is calculated by Eq.\ref{Eq5},
\begin{eqnarray}\label{Eq5}
  	C_i(x,y) = ||\phi_i^{1:M}(x,y)||_1
\end{eqnarray}

Then block-based average operator is utilized to calculate the final activity level map $\hat{C_i}$ by Eq.\ref{Eq6}.
\begin{eqnarray}\label{Eq6}
  	\hat{C_i}(x,y) = \frac{\sum_{a=-r}^{r}\sum_{b=-r}^{r}C_i(x+a,y+b)}{(2r+1)^2}
\end{eqnarray}
where $r$ determines the block size and in our strategy $r=1$.

After we get the final activity level map $\hat{C_i}$, $f^m$ is calculated by Eq.\ref{Eq7},
\begin{eqnarray}\label{Eq7}
  	&f^m(x,y) = \sum_{i=1}^k w_i(x,y) \times \phi_i^m(x,y), \\
	&w_i(x,y) = \frac{\hat{C_i}(x,y)}{\sum_{n=1}^{k}\hat{C_n}(x,y)} \nonumber
\end{eqnarray}

The final fused image will be reconstructed by decoder in which the fused feature maps $f^m$ as the input.

\begin{figure*}[ht]
\centering
\includegraphics[width=\linewidth]{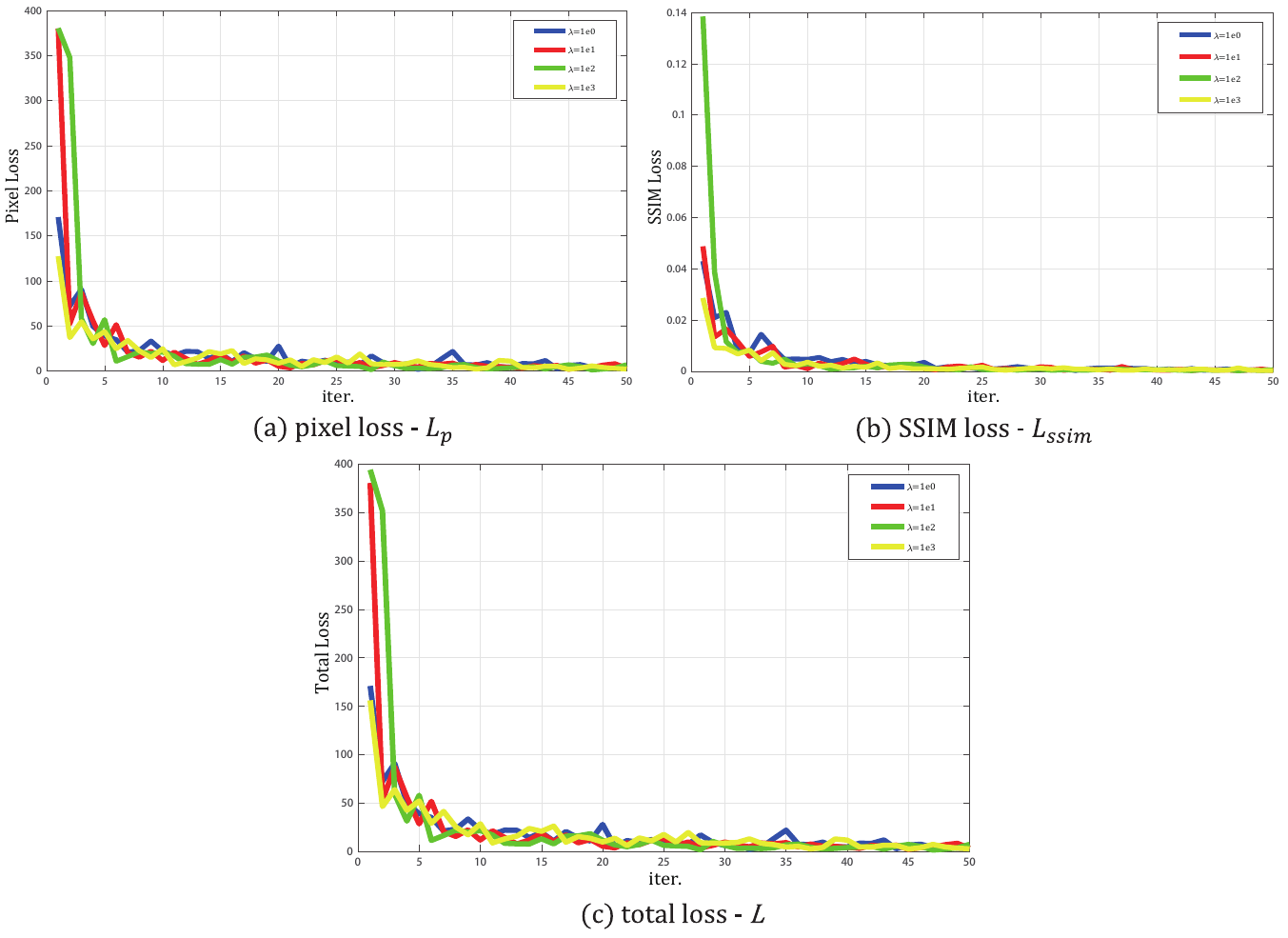}
\caption{The graph plot of pixel loss(a), SSIM loss(b) and total loss(c) in training step. Each point in horizontal axis indicates 100 iterations and we choose the first 5000 iterations. And ``blue'' - $\lambda=1$; ``red'' - $\lambda=10$; ``green'' - $\lambda=100$; ``yellow'' - $\lambda=1000$.}
\label{fig:lossplot}
\end{figure*}

\section{Experimental results and analysis}
\label{sec:experimrnt}

Firstly, we analyse the experiment of training phase. The graph plot of loss and validation are shown in Fig.\ref{fig:lossplot} and Fig.\ref{fig:valplot}.

And the purpose of testing experiment  is to validate the proposed fusion method using subjective and objective criteria and to carry out the comparison with existing methods. 

Then, the additional fusion results of infrared and visible(RGB) images are shown in Section \ref{sec:ivrgb}.

\subsection{Training phase analysis}
\label{sec:trainingana}

In training phase, we use MS-COCO\cite{18} as input images. In these source images, about 79000 images are utilized as input images, 1000 images are used to validate the reconstruction ability in every iteration.

As shown in Fig.\ref{fig:lossplot}, our network has rapid convergence with the increasing numerical index of SSIM loss weight $\lambda$ in the first 2000 iterations. As discussed in Section\ref{sec:training}, the orders of magnitude are different between pixel loss and SSIM loss. When the $\lambda$ increases, SSIM loss plays more important role in training phase.

\begin{figure*}[!ht]
\centering
\includegraphics[width=\linewidth]{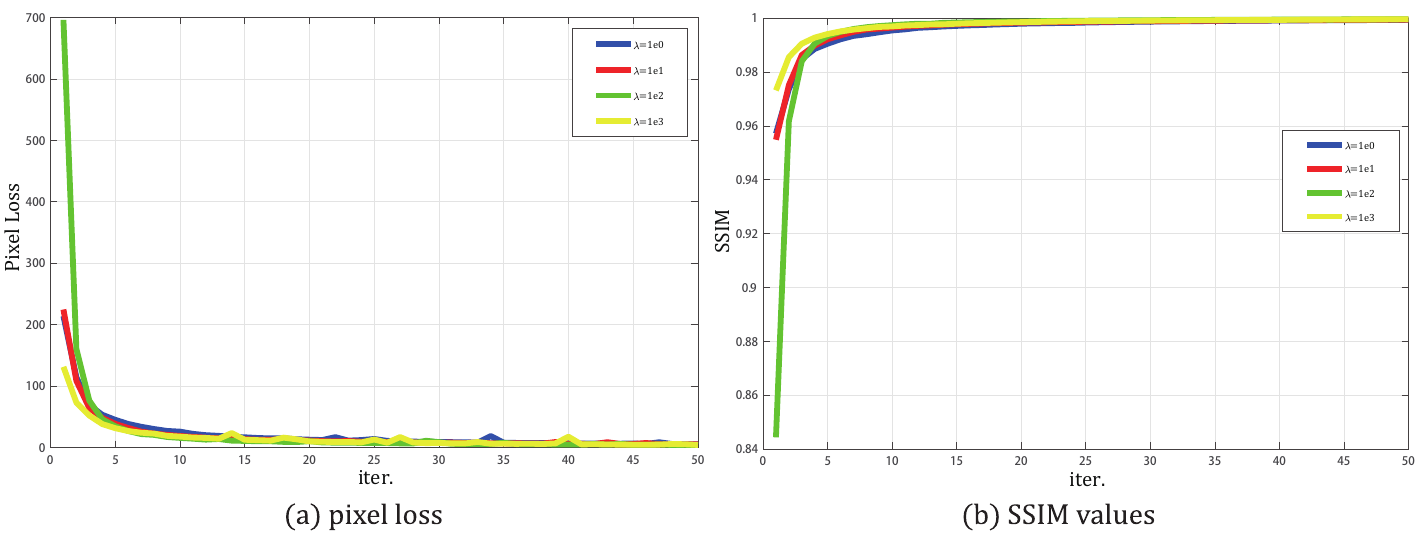}
\caption{The graph plot of pixel loss(a), SSIM values(b) in validation phase. Each point in horizontal axis indicates 100 iterations and we choose the first 5000 iterations. And ``blue'',  ``red'', ``green'', ``yellow'' indicate the SSIM loss weight $\lambda=1, 10, 100, 1000$, respectively.}
\label{fig:valplot}
\end{figure*}

In validation step, we choose 1000 images from MS-COCO as the inputs to our training network. The pixel loss and SSIM are utilized to evaluate the reconstruction ability.
From Fig.\ref{fig:valplot}, the validation graph plots indicate the SSIM loss plays important role with the increase of $\lambda$. With the iteration increase to 500, the pixel loss and SSIM achieve better values when $\lambda$ is set to larger values.

However, when iterations are larger than 40000, we will get the optimal weights, no matter which loss weights are chosen. In general, our network will get faster convergence with increase of $\lambda$ in early training phase. And the larger $\lambda$ will reduce the time consumption in training phase.

\subsection{Experimental Settings}
\begin{figure*}[!ht]
\centering
\includegraphics[width=\linewidth]{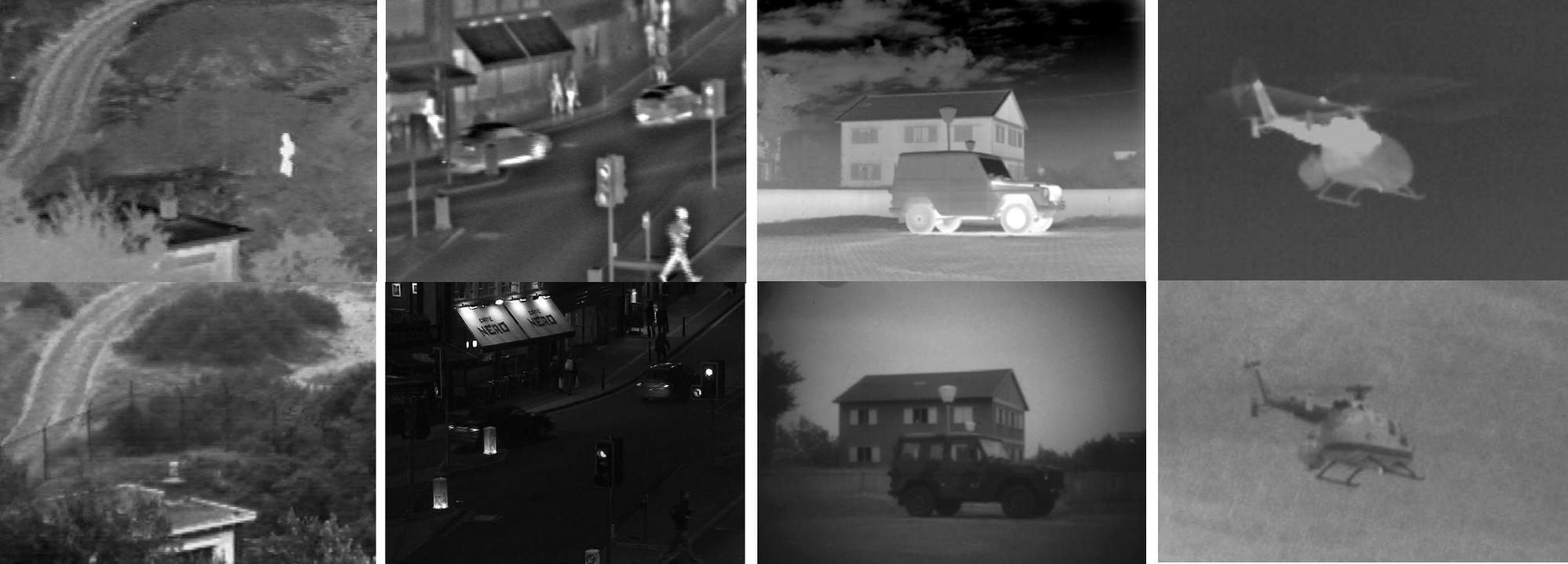}
\caption{Four pairs of source images. The top row contains infrared images, and the second row contains visible images.}
\label{fig:example}
\end{figure*}

In our experiment, the number of input images($k$) is 2 and the source infrared and visible images were collected from \cite{19} and \cite{20}. There are 20 pairs of our source images for the experiment and infrared and visible images are available at \cite{21}. A sample of these images is shown in Fig.\ref{fig:example}.

\begin{figure*}[!ht]
\centering
\includegraphics[width=\linewidth]{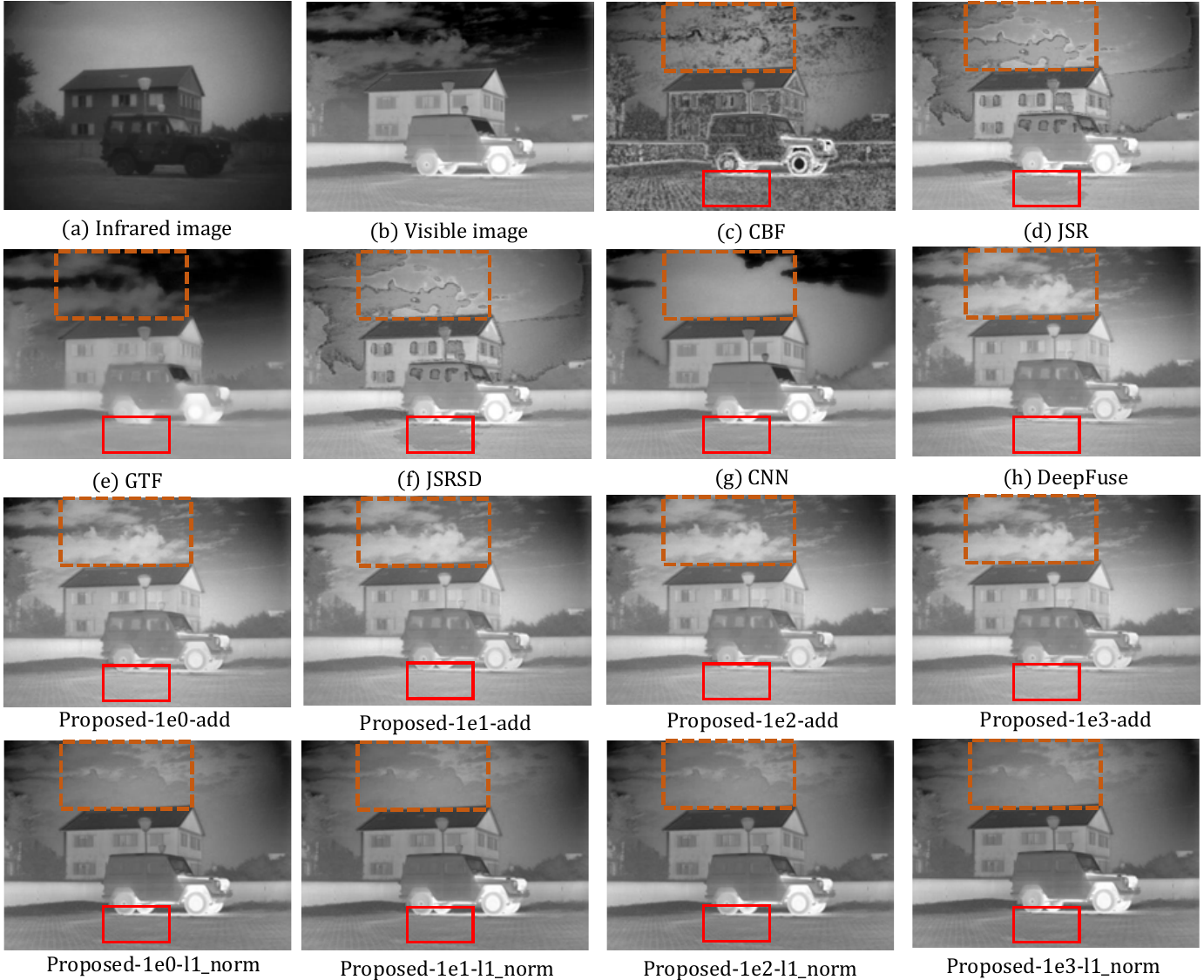}
\caption{Experiment on ``car'' images. (a) Infrared image; (b) Visible image; (c) DCHWT; (d) JSR; (e) GTF; (f) JSRSD; (g) WLS. (h) DeepFuse; The last two rows contain the fused images which obtained by proposed method with different SSIM weights and fusion strategy.}
\label{fig:car}
\end{figure*}

We compare the proposed method with several typical fusion methods, including cross bilateral filter fusion method(CBF)\cite{22}, the joint-sparse representation model(JSR)\cite{9}, gradient transfer and total variation minimization(GTF)\cite{23}, the JSR model with saliency detection fusion method(JSRSD)\cite{24}, deep convolutional neural network-based method(CNN)\cite{13} and the DeepFuse method(DeepFuse)\cite{15}. In our experiment, the filter size is set as $3\times3$ for DeepFuse methods.

For the purpose of quantitative comparison between our fusion method and other existing algorithms, seven quality metrics are utilized. They are: entropy(En); Qabf\cite{25}; the sum of the correlations of differences(SCD)\cite{26}; $FMI_w$ and $FMI_{dct}$\cite{27} which calculates mutual information (FMI) for the wavelet and discrete cosine features, respectively; modified structural similarity for no-reference image($SSIM_a$); and a new no-reference image fusion performance measure(MS\_SSIM)\cite{28}.

In our experiment, the $SSIM_a$ is calculated by Eq.\ref{Eq8},
\begin{eqnarray}\label{Eq8}
  	SSIM_a(F) = (SSIM(F,I_1)+SSIM(F,I_2))\times 0.5
\end{eqnarray}
where $SSIM(\cdot)$ denotes the structural similarity operation\cite{17}, $F$ is the fused image, and $I_1$, $I_2$ are source images. The value of $SSIM_a$ represents the ability to preserve structural information.

The fusion performance improves with the increasing numerical index of all these seven metrics.

\begin{figure*}[!ht]
\centering
\includegraphics[width=\linewidth]{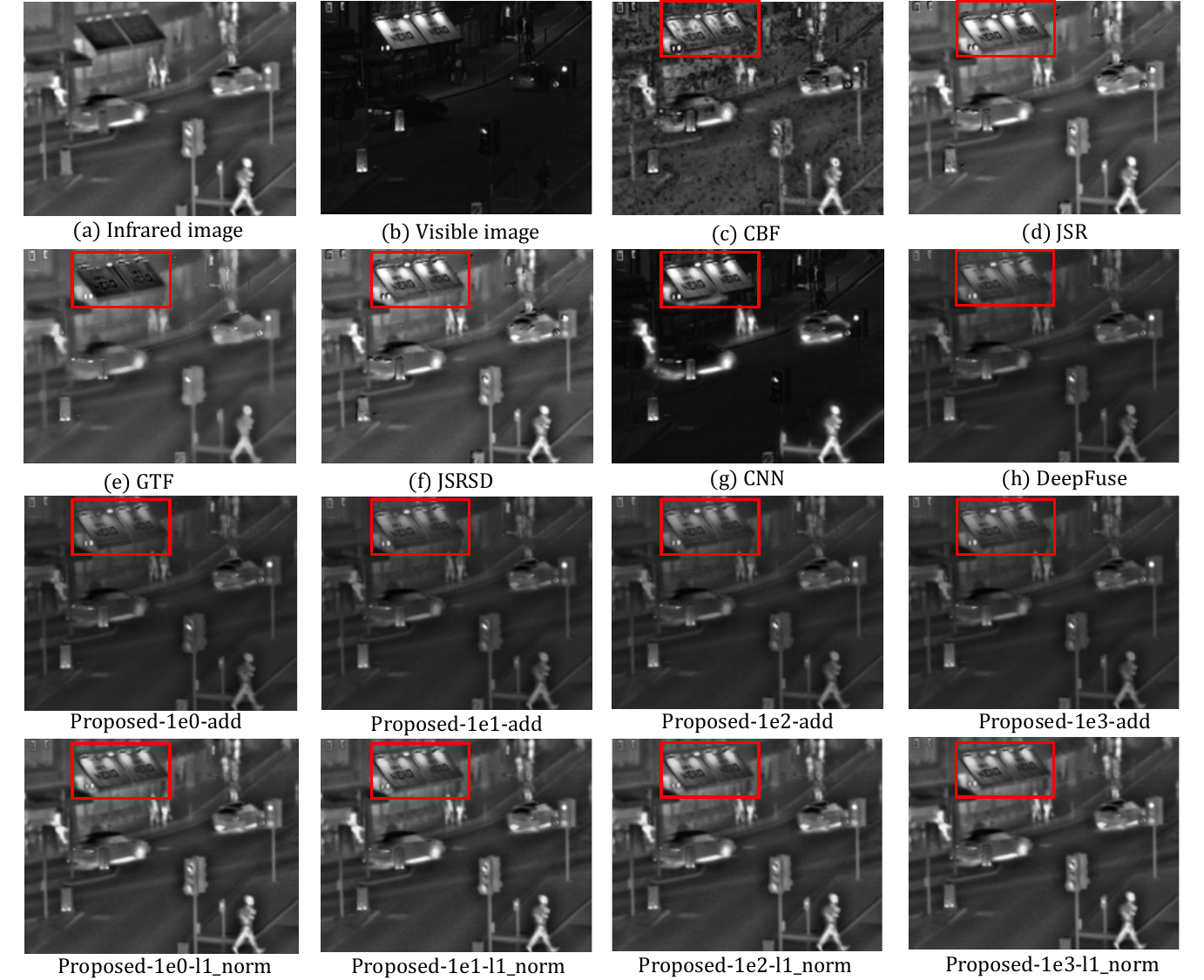}
\caption{Experiment on ``street'' images. (a) Infrared image; (b) Visible image; (c) DCHWT; (d) JSR; (e) GTF; (f) JSRSD; (g) WLS. (h) DeepFuse; The last two rows contain the fused images which obtained by proposed method with different SSIM weights and fusion strategy.}
\label{fig:street}
\end{figure*}

\subsection{Fusion methods Evaluation}

The fused images obtained by the six existing methods and the proposed method use different parameters which are shown in Fig.\ref{fig:car} and Fig.\ref{fig:street}. Due to the space limit, we evaluate the relative performance of the fusion methods on two pairs of images(``car'' and ``street'').

The fused images obtained by CBF, JSR and JSRSD have more artificial noise and the saliency features are not clear, such as sky(orange and dotted) and floor(red and solid) in Fig.\ref{fig:car} and billboard(red box) in Fig.\ref{fig:street}.

On the other hand, the fused images obtained by the proposed method contain less noise in the red box no matter what parameters were chosen. Compared with GTF, CNN and DeepFuse, our fusion method preserves more detailed information in red box, as we can see from Fig.\ref{fig:car}.

In Fig.\ref{fig:street}, the fused image is darker than other images when the CNN-based method is utilized to fuse images. The reason of this phenomenon is probably CNN-based method is not suitable for infrared and visible images. On the contrary, the fused images obtained by our method look more natural.

However, as there is almost no visiual difference between DeepFuse and proposed method in human sensitivity, we choose several objective metrics to evaluate the fusion performance next.

The average values of seven metrics for 20 fused images which are obtained by existing methods and the proposed fusion method are shown in Table \ref{tab:values}.

The best values for quality metrics are indicated in \textbf{bold} and the second-best values are indicated in \emph{\color{blue}{blue and italic}}. As we can see, the proposed method which adopts addition and $l_1$-norm strategies has five best average values (En, Qabf, SCD, $FMI_{dct}$, $SSIM_a$) and two second-best values ($FMI_w$, MS\_SSIM). 

Our method has best values in $FMI_{dct}$, $SSIM_a$, this denotes that our method preserves more structural information and features. The fused images obtained by proposed method are more natural and contain less artificial noise because of the best values of En, Qabf and SCD.

With different fusion strategy (addition and $l_1$-norm) utilized in to our network, our algorithm still has best or second-best values in seven quality metrics. This means our network is an effective architecture for infrared and visible image fusion task.

\begin{table*}[ht]
\centering
\caption{\label{tab:values}The average values of quality metrics for 20 fused images. \emph{\textbf{Addition}} and \emph{\textbf{$l_1$-norm}} denote the fusion strategies which we used in our method; \emph{\textbf{Densefuse\_1e0}} - \emph{\textbf{Densefuse\_1e3}} indicate the different SSIM loss weights($\lambda$).}
\resizebox{\textwidth}{!}{
\begin{tabular}{|c|c|c|c|c|c|c|c|c|c|c|}
\hline 
\multicolumn{3}{|c|}{Methods}	& En		& Qabf\cite{25} & SCD\cite{26}  & $FMI_w$\cite{27} & $FMI_{dct}$\cite{27} & $SSIM_a$ & MS\_SSIM\cite{28} \\
\hline
\multicolumn{3}{|c|}{CBF\cite{22}}		&6.81494	&0.44119	&1.38963	&0.32012	&0.26619	&0.60304	&0.70879\\
\hline
\multicolumn{3}{|c|}{JSR\cite{9}}		&6.78576	&0.32572	&1.59136	&0.18506	&0.14184	&0.53906	&0.75523\\
\hline
\multicolumn{3}{|c|}{GTF\cite{23}}		&6.63597	&0.40992	&1.00488	&0.41004	&0.39384	&0.70369	&0.80844\\
\hline
\multicolumn{3}{|c|}{JSRSD\cite{24}}		&6.78441	&0.32553	&1.59124	&0.18502	&0.14201	&0.53963	&0.75517\\
\hline
\multicolumn{3}{|c|}{CNN\cite{13}}		&6.80593	&0.29451	&1.48060	&\textbf{0.53954}	&0.35746	&0.71109	&0.80772\\
\hline
\multicolumn{3}{|c|}{DeepFuse\cite{15}}	&\emph{\color{blue}{6.68170}}	&0.43989	&\emph{\color{blue}{1.84525}}	&0.42438	&\emph{\color{blue}{0.41357}}	&\emph{\color{blue}{0.72949}}	&\textbf{0.93353}\\
\hline
\multirow{8}*{ours} &
\multirow{4}*{Addition} &
  $\lambda=1e0$	&6.66280	&0.44114	&\textbf{1.84929}	&\emph{\color{blue}{0.42713}}	&\textbf{0.41557}	&\textbf{0.73159}	&\emph{\color{blue}{0.93039}}\\
\cline{3-10}
&&$\lambda=1e1$	&6.65139	&0.44039	&\textbf{1.84549}	&\emph{\color{blue}{0.42707}}	&\textbf{0.41552}	&\textbf{0.73246}	&\emph{\color{blue}{0.92896}}\\
\cline{3-10}
&&$\lambda=1e2$	&6.65426	&\emph{\color{blue}{0.44190}}	&\textbf{1.84854}	&\emph{\color{blue}{0.42731}}	&\textbf{0.41587}	&\textbf{0.73186}	&\emph{\color{blue}{0.92995}}\\
\cline{3-10}
&&$\lambda=1e3$	&6.64377	&0.43831	&1.84172	&\emph{\color{blue}{0.42699}}	&\textbf{0.41558}	&\textbf{0.73259}	&\emph{\color{blue}{0.92794}}\\
\cline{2-10}
&\multirow{4}*{$l_1$-norm} &
  $\lambda=1e0$	&\textbf{6.83278}	&\textbf{0.47560}	&1.71182		&\emph{\color{blue}{0.43191}}	&0.38062	&0.71880	&0.85707\\
\cline{3-10}
&&$\lambda=1e1$	&6.81348			&\textbf{0.47680}	&1.71264		&\emph{\color{blue}{0.43224}}	&0.38048	&0.72052	&0.85803\\
\cline{3-10}
&&$\lambda=1e2$	&\textbf{6.83091}	&\textbf{0.47684}	&1.71705		&\emph{\color{blue}{0.43129}}	&0.38109	&0.71901	&0.85975\\
\cline{3-10}
&&$\lambda=1e3$	&\textbf{6.84189}	&\textbf{0.47595}	&1.72000		&\emph{\color{blue}{0.43147}}	&0.38404	&0.72106	&0.86340\\
\cline{3-10}
\hline
\end{tabular}}
\end{table*}

\subsection{Additional results for RGB images and infrared images}
\label{sec:ivrgb}

Apart from the gray scale images fusion task, our fusion algorithm can be used to fuse the visible images which include RGB channels and infrared images. The input images are collected from \cite{34}.

\begin{figure}[!ht]
\centering
\includegraphics[width=\linewidth]{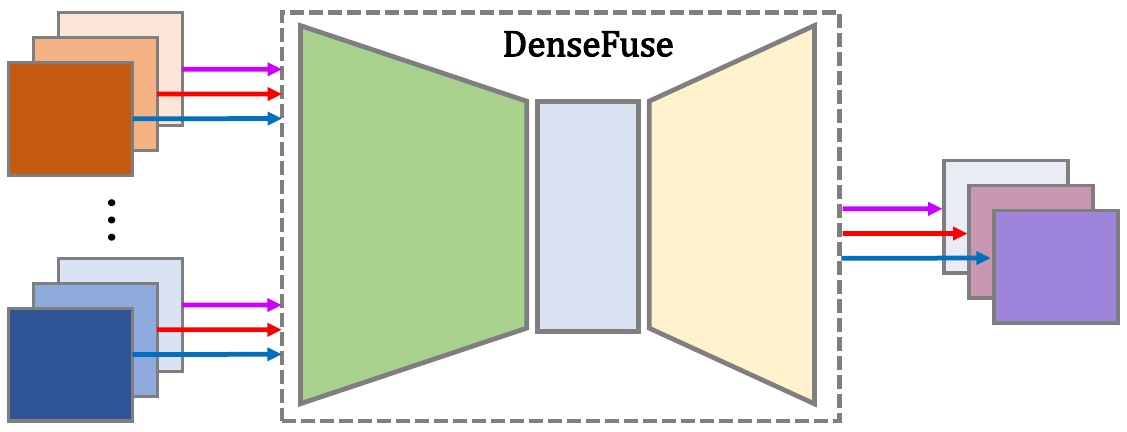}
\caption{The fusion framework for RGB images.}
\label{fig:frgb}
\end{figure}

As shown in Fig.\ref{fig:frgb}. We use the fixed network(encoder, fusion layer, decoder) to fuse these images. When the RGB scale images are processed, each channel in RGB is treated as one gray scale image. Therefore there are three pairs of channel for a pair of RGB scale images, each pair of channel will be the input images to our network. Then, three fused channels are obtained by our network and we combine these fused channels as one fused RGB images.

\begin{figure*}[!ht]
\centering
\includegraphics[width=\linewidth]{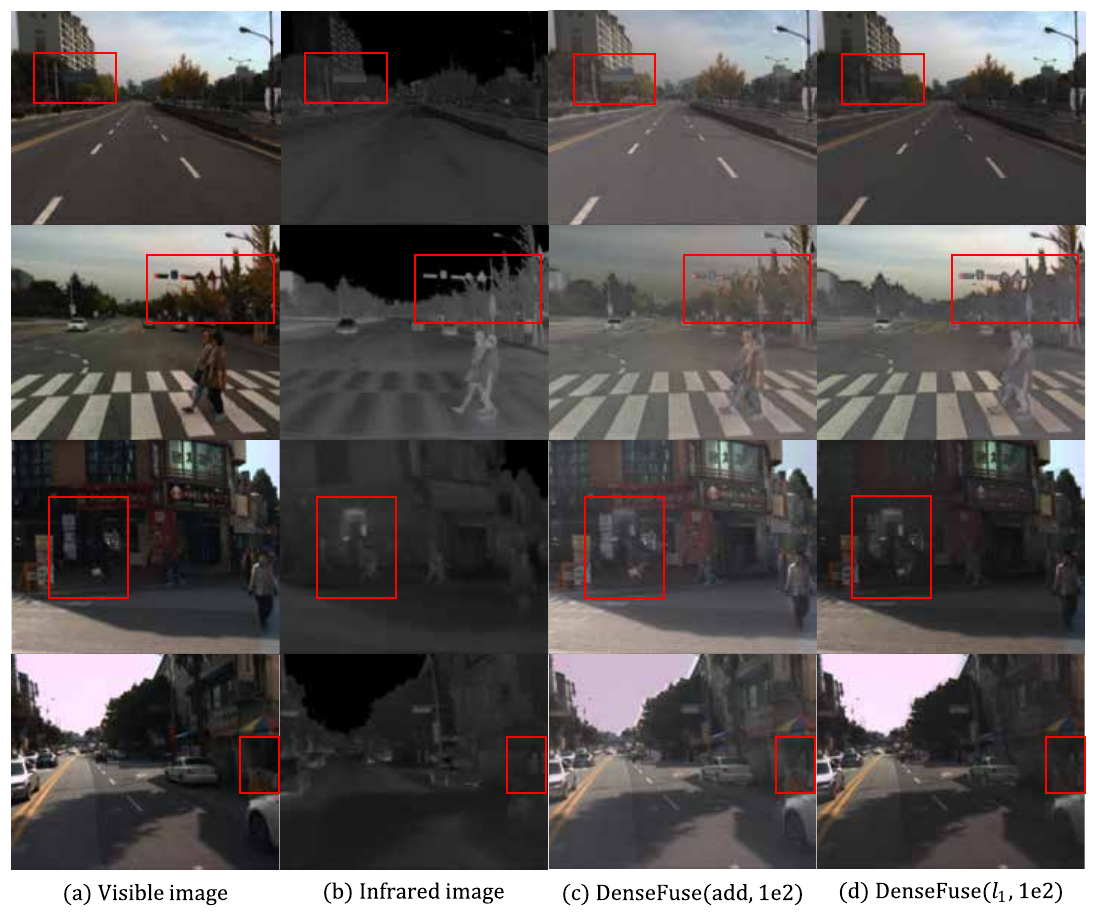}
\caption{Fused results for infrared and visible(RGB) images with DenseFuse. (a) Visible image; (b) Infrared image; (c) Addition strategy and loss weight(1e2); (d) $l_1$-norm strategy and loss weight(1e2).}
\label{fig:ivrgb}
\end{figure*}

The fused results for RGB scale images and infrared images are shown in Fig.\ref{fig:ivrgb}. The fusion results of other images and multi-focus images are available at our supplementary material.

\section{Conclusion}
\label{sec:con}
In this paper, we present a novel and effective deep learning architecture based on CNN and dense block for infrared and visible image fusion problem. And this algorithm is not only used to fuse gray scale images, but also can be applied to fuse RGB scale images. 

Our network has three parts: encoder, fusion layer and decoder. Firstly, the source images (infrared and visible images) are utilized as the input of encoder. And the features maps are obtained by CNN layer and dense block, which are fused by fusion strategy (addition and $l_1$-norm). After fusion layer, the feature maps are integrated into one feature map which contains all salient features from source images. Finally, the fused image is reconstructed by decoder network. We use both subjective and objective quality metrics to evaluate our fusion method. The experimental results show that the proposed method exhibits state-of-the-art fusion performance.

Initial experiments show that our network architecture can be applied to other image fusion problems with appropriate fusion layer, such as multi-focus image fusion, multi-exposure image fusion and medical image fusion.


\begin{IEEEbiography}[{\includegraphics[width=1in,height=1.25in,clip,keepaspectratio]{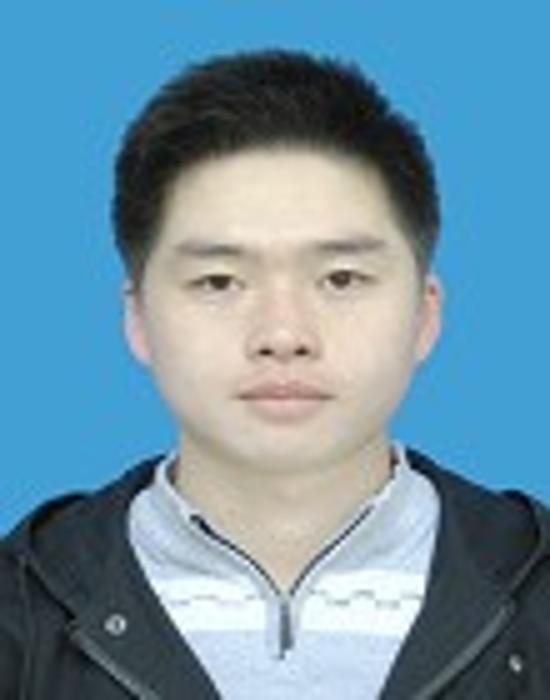}}]{Hui Li}
received the B.E. degree in School of Internet of Things Engineering from Jiangnan University, China. He is currently a PhD student in the Jiangsu Provincial Engineerinig Laboratory of Pattern Recognition and Computational Intelligence, Jiangnan University. His research interests include image fusion, machine learning and deep learning.
\end{IEEEbiography}

\begin{IEEEbiography}[{\includegraphics[width=1in,height=1.25in,clip,keepaspectratio]{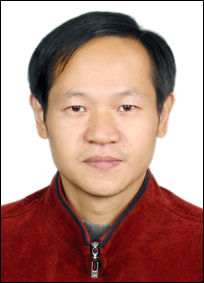}}]{Xiao-Jun Wu}
received the B.S. degree in mathematics from Nanjing Normal University, Nanjing, China,
in 1991, and the M.S. degree and Ph.D. degree
in pattern recognition and intelligent system from
the Nanjing University of Science and Technology, Nanjing, in 1996 and 2002, respectively. From
1996 to 2006, he taught at the School of Electronics
and Information, Jiangsu University of Science and
Technology, where he was promoted to Professor.
He has been with the School of Information Engineering, Jiangnan University, since 2006, where he
is a Professor of Computer Science and Technology. He was a Visiting
Researcher with the Centre for Vision, Speech, and Signal Processing, University of Surrey, U.K., from 2003 to 2004. He has published over 200 papers.
His current research interests include pattern recognition, computer vision, and
computational intelligence. He was a Fellow of the International Institute for
Software Technology, United Nations University, from 1999 to 2000. He was
a recipient of the Most Outstanding Postgraduate Award from the Nanjing
University of Science and Technology.
\end{IEEEbiography}

\vfill

\end{document}